\documentclass[letterpaper, 10 pt, conference]{ieeeconf}  % Comment this line out if you need a4paper

\IEEEoverridecommandlockouts                              % This command is only needed if 
\usepackage{graphicx}
\usepackage{subcaption}
\usepackage[export]{adjustbox}
\usepackage{amssymb}
\usepackage{amsmath}
\usepackage{commath}
\usepackage{mathtools}
\usepackage{booktabs}
\usepackage[table,xcdraw]{xcolor}
\usepackage{hyperref} % For clickable links

\title{\LARGE \bf
Physics-Informed Neural Networks with Unscented Kalman Filter for Sensorless Joint Torque Estimation in Humanoid Robots
}

\author{Ines Sorrentino$^{1,2}$, Giulio Romualdi$^{1}$, Lorenzo Moretti$^{1}$, Silvio Traversaro$^{1}$, Daniele Pucci$^{1,2}$% <-this % stops a space
\thanks{$^{1} $Ines Sorrentino and Daniele Pucci are with  the
Artificial and Mechanical Intelligence, Istituto Italiano di Tecnologia, 16163
Genoa, Italy, and also with the Department of Computer Science, The University
of Manchester, Manchester M13 9PL, U.K. (e-mail: ines.sorrentino@iit.it; daniele.pucci@iit.it).}
\thanks{$^{2} $Giulio Romualdi, Lorenzo Moretti, and Silvio Traversaro are with the Artificial and Mechanical Intelligence, Istituto Italiano di Tecnologia, 16163 Genoa, Italy (e-mail: giulio.romualdi@iit.it; lorenzo.moretti@iit.it; silvio.traversaro@iit.it).}%
}

\begin{document}

\maketitle

%%%%%%%%%%%%%%%%%%%%%%%%%%%%%%%%%%%%%%%%%%%%%%%%%%%%%%%%%%%%%%%%%%%%%%%%%%%%%%%%
\begin{abstract}

This paper presents a novel framework for whole-body torque control of humanoid robots without joint torque sensors, designed for systems with electric motors and high-ratio harmonic drives. The approach integrates Physics-Informed Neural Networks (PINNs) for friction modeling and Unscented Kalman Filtering (UKF) for joint torque estimation, within a real-time torque control architecture. PINNs estimate nonlinear static and dynamic friction from joint and motor velocity readings, capturing effects like motor actuation without joint movement. The UKF utilizes PINN-based friction estimates as direct measurement inputs, improving torque estimation robustness. Experimental validation on the ergoCub humanoid robot demonstrates improved torque tracking accuracy, enhanced energy efficiency, and superior disturbance rejection compared to the state-of-the-art Recursive Newton-Euler Algorithm (RNEA), using a dynamic balancing experiment. The framework’s scalability is shown by consistent performance across robots with similar hardware but different friction characteristics, without re-identification. Furthermore, a comparative analysis with position control highlights the advantages of the proposed torque control approach. The results establish the method as a scalable and practical solution for sensorless torque control in humanoid robots, ensuring torque tracking, adaptability, and stability in dynamic environments.

\end{abstract}

%%%%%%%%%%%%%%%%%%%%%%%%%%%%%%%%%%%%%%%%%%

\section{Introduction}
\label{sec:introdcution}

The ongoing evolution of humanoid robotics is expanding the boundaries of functionality, safety, and integration into human environments. Humanoid robots are designed for tasks like locomotion, manipulation, and human-robot interaction, demanding safe adaptability, compliant interaction, and precision, which can be facilitated by introducing compliance in the robot's joints~\cite{dafarra_icub3_2024,romualdi_online_2024,darvish2023teleoperation}. Joint torque control is widely adopted among various control strategies as it enables compliance and adaptability. However, it typically relies on direct torque measurements, often impractical due to cost, integration complexity, and sensor limitations~\cite{dantec_whole-body_2022,singh_learning_2023,mesesan2019dynamic}.
As a result, many humanoid robots without joint torque sensors rely on rigid control strategies that use predefined joint trajectories. While such approaches are effective in controlled environments, they often fail in dynamic, real-world scenarios where precise torque estimation is crucial for robust interaction with unpredictable surroundings. To address the limitations of current sensorless torque control methods, this paper introduces a novel framework that synergistically integrates Physics-Informed Neural Networks (PINNs) for complex friction modeling and an Unscented Kalman Filter (UKF) for robust joint torque estimation.

\begin{figure}[t]
    \centering
    \adjustbox{width=\linewidth}{%
        \includegraphics{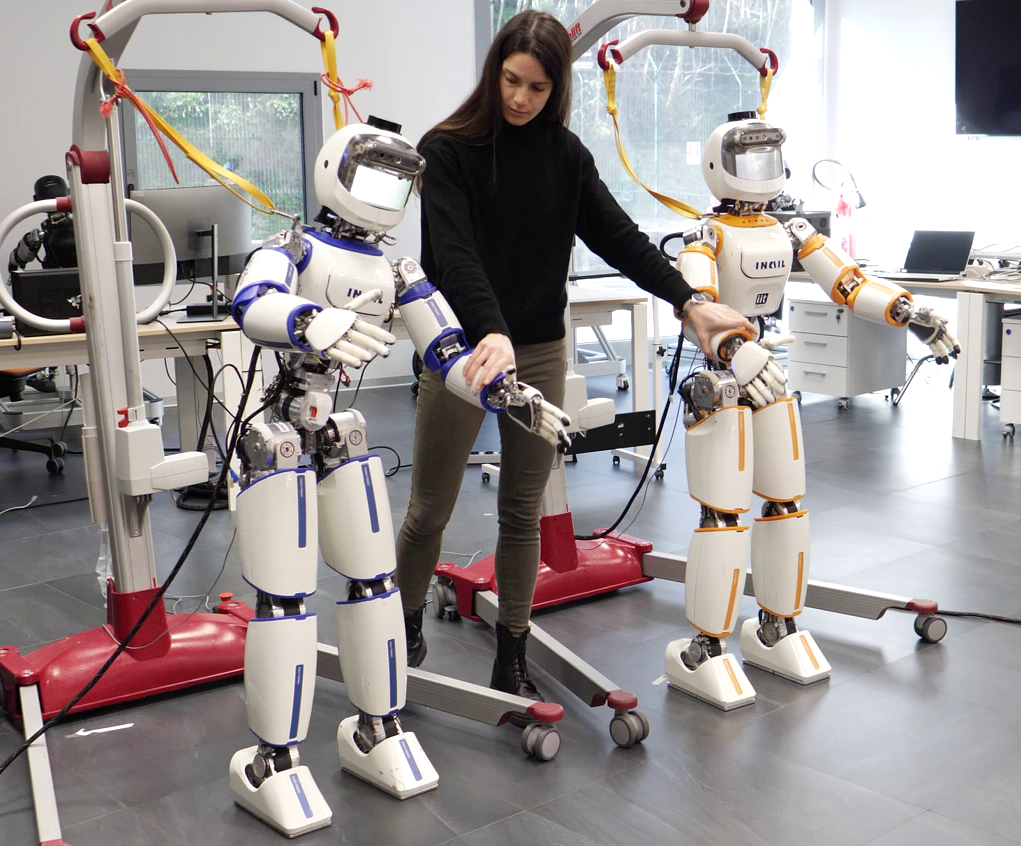}
    }
    \caption{Two ergoCub robots performing the balancing task during dynamic interactions.}
    \label{fig:ergocub}
    \vspace{-18pt}
\end{figure}

Existing methods estimate joint torques in real-time by combining sensor data (e.g., motor currents, joint positions, and velocities) with advanced algorithms~\cite{sorrentino_ukf-based_2024,del_prete_implementing_2016,fumagalli2012force}. A common model-based approach is the Recursive Newton-Euler Algorithm (RNEA)~\cite{nori2015icub,del_prete_implementing_2016,fumagalli2012force}. RNEA estimates joint torques based on rigid-body dynamics but is prone to inaccuracies from unmodeled dynamics, including friction, gearbox elasticity, sensor noise, and unmeasured external contacts~\cite{sorrentino_ukf-based_2024,zhang2015torque,nava2018exploiting}. Alternatively, some solutions compute joint torques from motor current and encoder data. While these offer fast response, they often assume rigid gear transmission, neglecting elastic deformations and backlash in high-ratio gears~\cite{nagamatsu2017distributed}. Sensorless variable impedance control is another approach, dynamically adjusting joint stiffness and damping based on interaction forces estimated from motor currents and encoders. Still, its accuracy heavily relies on precise robot dynamics models~\cite{liu2022sensorless}.

 These torque estimation methods are all limited by complex friction dynamics, which sensorless control frameworks commonly address through friction estimation and compensation strategies. While basic models like Coulomb and viscous friction offer limited accuracy~\cite{del_prete_implementing_2016,nagamatsu2017distributed}, advanced models such as LuGre and Stribeck capture more intricate friction effects~\cite{awrejcewicz2005analysis,jin2019joint}. Recent advancements have integrated physics-based modeling with data-driven techniques to estimate friction dynamics under varying loads. While these hybrid approaches leverage physical insights to enhance data-driven models, their complexity and the requirements for specialized setups for accurate friction identification can limit their applicability. It is important to distinguish this application of physics-informed modeling for friction estimation from the broader use of Physics-Informed Neural Networks (PINNs) in robotics. PINNs have gained traction in diverse areas, including dynamic modeling~\cite{song2025study}, parameter identification in collaborative robots~\cite{yang2023collaborative}, and hybrid modeling for robot control~\cite{liu2024physics}, demonstrating their versatility in embedding physical laws within neural network architectures. However, their application for addressing complex friction dynamics in floating-base humanoid robots remains largely unexplored. Existing studies often focus on offline modeling or parameter identification rather than real-time compensation for frictional disturbances.

To address the limitations of existing sensorless torque control methods, this paper builds upon our previous work in UKF-based torque estimation~\cite{sorrentino_ukf-based_2024} and PINN-based friction modeling~\cite{sorrentino2024physics} by presenting a novel integrated framework for whole-body torque control of humanoid robots. Key advancements include:
\begin{enumerate}
\item A significant enhancement of our PINN-based friction modeling, utilizing motor and joint velocity buffers as inputs to capture static friction and extending the modeling from two joints to all leg joints.
\item An extension of our UKF formulation from fixed-base to floating-base systems for real-time adaptation to dynamic conditions. Furthermore, we now introduce PINN-predicted friction torques as direct measurements within the UKF, enhancing estimation accuracy and robustness.
\item The synergistic combination of these advanced estimation techniques within a unified real-time control architecture.
\item The framework achieves consistent performance across similar robotic platforms without re-identification.
\end{enumerate}
This work makes significant contributions: bridging critical gaps in sensorless torque control and setting a new benchmark for achieving compliant, robust, and energy-efficient control in humanoid robots. The effectiveness of the proposed framework has been demonstrated through rigorous experimental validation on the ergoCub humanoid robot. Key results include substantial advancements, such as reduced torque tracking RMSE and improved energy efficiency compared to RNEA. Furthermore, the framework exhibits superior disturbance rejection and adaptability, even in dynamic scenarios. These results also include a comparative analysis with position control, highlighting the advantages of the proposed torque control approach. Indeed, the compliance achieved through our solution, is essential for robots to operate effectively and safely in real-world scenarios, allowing them to adapt to unpredictable forces and maintain stability.

The remainder of the paper is organized as follows. Section~\ref{sec:background} reviews the enabling technologies and fundamental principles for torque estimation and friction modeling. Section~\ref{sec:methodology} details the torque control framework, while Section~\ref{sec:results} presents experimental validations, including comparisons with existing methods. Finally, Section~\ref{sec:conclusions} concludes the paper.
\looseness=-1

\section{Background}
\label{sec:background}

\subsection{Notation}
\begin{itemize}
\item $I_{m \times n}$ and $0_{m \times n}$ denote the $m \times n$ identity and zero matrices respectively; when $m=n$ we use $I_{n}$ and $0_{n}$.
\item $A$ denotes the inertial frame.
\item $\prescript{A}{}{p}_B$ is a vector connecting the origin of frame $A$ and the origin of frame $B$ expressed in frame $A$.
\item Given $\prescript{A}{}{p}_B$ and $\prescript{B}{}{p}_C$,  $\prescript{A}{}{p}_C = \prescript{A}{}{R}_B \prescript{B}{}{p}_C + \prescript{A}{}{p}_B= \prescript{A}{}{H}_B \left[
  \prescript{B}{}{p}_C ^\top , \;1
\right]^\top$. $\prescript{A}{}{H}_B$ is the homogeneous transformation and $\prescript{A}{}{R}_B \in SO(3)$ is the rotation matrix.
\item $\scalebox{0.8}{$\prescript{B}{}{\textrm{v}}_{A,B} = \left[ {}^B {v}_{A,B} ^\top , {}^B {\omega}_{A,B}^\top \right]^\top \in \mathbb{R}^6$}$ is the velocity of the frame $B$ with respect to the frame $A$ expressed in $B$.
\item $\times$ is the cross product in $\mathbb{R}^{3}$.
\item ${}^B g$ is the gravity vector expressed in $B$.
\item $\scalebox{0.8}{$\alpha_{A,B}^g := \alpha_{A,B} - \begin{bmatrix} ({}^B R_A {}^A g)^\top & 0_{1 \times 3} \end{bmatrix}^\top$}$ is the proper acceleration of the frame $B$ with respect to $A$.
\item  $\scalebox{0.8}{${\mathrm{f}}_B \in \mathbb{R}^6$}$ is the wrench acting on a point of a rigid body expressed in the frame $B$.
\item $\hat{}$ symbol identifies variables estimated through the UKF.
\end{itemize}
\looseness=-1

\subsection{Humanoid Robot Model}
A humanoid robot is a floating-base multi-body system described by the configuration $q = (\prescript{A}{}{p}_B, \prescript{A}{}{R}_B, s) \in  \mathbb{R}^3 \times SO(3) \times \mathbb{R}^n$, where $s \in \mathbb{R}^n$ represents the $n$ joint positions, and $B$ denotes the base frame. The velocity of the system is given by the set $\nu = (\prescript{A}{}{\dot{p}}_B, \prescript{A}{}{\omega}_B, \dot{s}) \in \mathbb{R}^{6+n}$. The equation of motion of a multi-body system in terms of proper acceleration~\cite{traversaro_modelling_nodate,featherstone_rigid_2014} is described by
\begin{equation}\label{eq:dynamics_sensor_frame}\small
\vspace{-2pt}
    {M(q) \begin{bmatrix} {{}^B \alpha^g_{A,B}} \\ \ddot{s} \end{bmatrix}} {+} {C(q, {{}^B \nu_{A,B}})} {=} B \tau {+} {\sum_{k} {J_k^\top(q) f_{ext}^k} },
    \vspace{-2pt}
\end{equation}
where \mbox{$M = \begin{bmatrix} M_b & M_{bs} \\ M_{sb} & M_s \end{bmatrix} \in \mathbb{R}^{(6+n) \times (6+n)}$} is the mass matrix, \mbox{$h(q, \nu) \in \mathbb{R}^{6+n}$} accounts for the Coriolis, centrifugal and gravitational effects, $\tau \in \mathbb{R}^{n}$ are the joint torques, $B \in \mathbb{R}^{(6+n) \times n}$ is the selection matrix, $f_{ext}^k \in \mathbb{R}^6$ is the $k$-th contact wrench expressed in the contact frame, $J_k(q)$ is the Jacobian associated to the $k$-th contact wrench.

\vspace{-2pt}
\subsection{Motor and Gearbox Dynamics}
The dynamics of electric motors mounted with high-ratio gearboxes can be described by the relationship:
\begin{equation}
\label{eq:motor_gearbox_dynamics}
\vspace{-2pt}
    k_{t} I_m = J_m \ddot{\theta} + \frac{1}{R} \tau_F + \frac{1}{R} \tau ,
    \vspace{-2pt}
\end{equation}
where $k_t$ is the torque constant, $I_m$ is the motor current, $J_m$ is the rotational inertia of the motor shaft, $\ddot{\theta}$ is the motor shaft angular acceleration, $\tau_F$ is the friction torque, $\tau$ is the torque to the load, and $R$ is the reduction ratio~\cite{taghirad_experimental_1996}. In systems mounting high-ratio gearboxes and low angular accelerations operating conditions $J_m \ddot{\theta} \ll \frac{1}{R} \tau_F$.

\section{Methodology}
\label{sec:methodology}

This section describes the estimation techniques developed for sensorless torque control of humanoid robots with high-ratio harmonic drives. The proposed approach combines Physics-Informed Neural Networks (PINNs) for advanced friction modeling and Unscented Kalman Filters (UKFs) for joint torque estimation.

\subsection{PINN Architecture}
\label{subsec:pinn}
This work extends previous applications of Physics-Informed Neural Networks (PINNs) for friction modeling~\cite{sorrentino2024physics}. In the previous approach to PINN-based friction modeling, friction estimation relied on buffers of differences between motor and joint positions and joint velocities. However, joint velocity estimation was noisy, and the choice of using differences between motor and joint positions was due to the absence of motor velocity information. Nevertheless, due to the lower resolution of the joint encoder, small discrepancies appeared even in the absence of actual movement. This work extends previous PINN applications for friction modeling by incorporating motor and joint velocity buffers as inputs to overcome previous limitations and enable the network to capture static friction when the motor is actuated accurately, but the joint remains stationary. We employ a custom Kalman Filter (KF) that provides smoother and more accurate velocity estimates, significantly improving the precision of the PINN-based friction modeling.
The KF operates with the following system dynamics:
\begin{equation}
\begin{bmatrix}
x_{k+1} \\ \dot{x}_{k+1} \\ \ddot{x}_{k+1}
\end{bmatrix} = 
\begin{bmatrix}
1 & \Delta T & \frac{ \Delta T^2}{2} \\ 0 & 1 & \Delta T \\ 0 & 0 & 1
\end{bmatrix}
\begin{bmatrix}
x_{k} \\ \dot{x}_{k} \\ \ddot{x}_{k}
\end{bmatrix} + v
\vspace{-2pt}
\end{equation}
The filter covariances are tuned to minimize the jerk and acceleration, ensuring smooth transitions, and align the estimated position with the measured position and the positions derived from velocity and acceleration integration.

The PINN predicts $\tau_{F}$ using a hybrid loss function combining data-driven and physics-based components:
\begin{equation}
\begin{aligned}
    \vspace{-2pt}
    \mathcal{L} = \mathcal{L}_{\text{data}} &{+} \mathcal{L}_{\text{physics}} =
    (1-\lambda) \frac{1}{N} \sum_{i=1}^{N} \left( {\tau}_{F,\text{pred}} {-} {\tau}_{F,\text{true}} \right)^2 \\
    & + \lambda \frac{1}{N} \sum_{i=1}^{N} \left( {\tau}_{F,\text{pred}} {-} {\tau}_{F,\text{physics}} \right)^2,
    \vspace{-2pt}
\end{aligned}
\end{equation}
where $\lambda \in [0, 1]$ balances the contributions of the data and physics losses, and ${\tau}_{F,\text{physics}}$ is derived from the Stribeck-Coulomb-Viscous (SCV) friction model~\cite{sorrentino2024physics}.

The PINN architecture features two hidden layers with ReLU activation functions, dropout regularization to prevent overfitting to training data, and a linear output layer. The hyperparameters for the PINN are optimized using the Weights and Biases platform~\cite{biewald2020experiment}, which employs a Random Search algorithm~\cite{bergstra2012random} to efficiently explore the hyperparameter space while balancing computational cost and training efficiency.

\begin{figure*}[ht]
    \centering
    \vspace{2pt}
    \includegraphics[width=460pt]{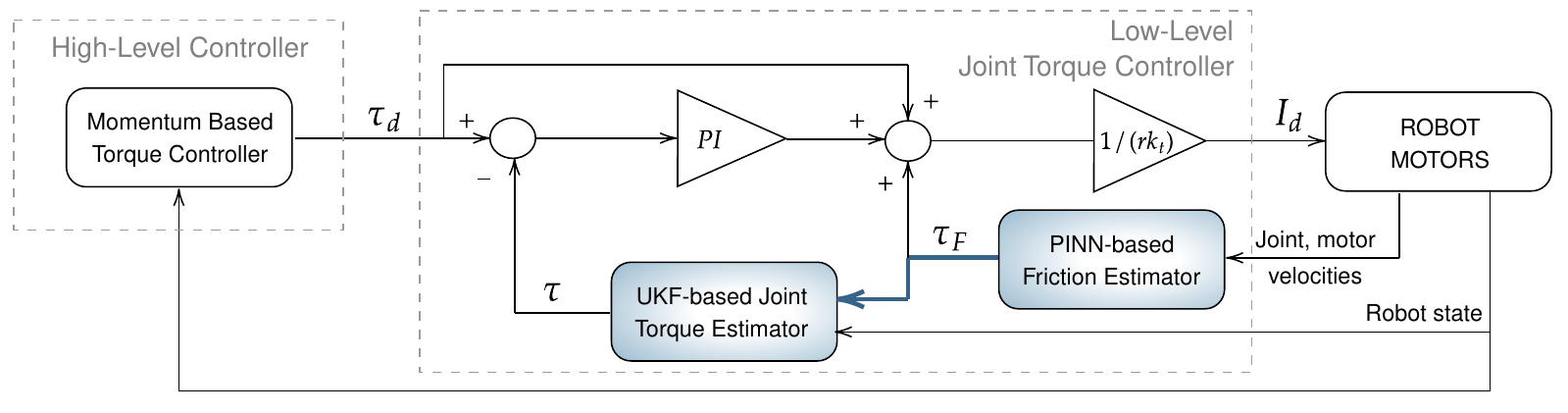}
    % \vspace{-6pt}
    \caption{Block diagram of the multi-layer torque control architecture implemented on the ergoCub humanoid robot. The high-level controller computes the desired joint torques ($\tau_d$) based on~\cite{nava_stability_nodate}. The low-level controller combines the desired torques $\tau_d$, the feedback ($\tau$) from the UKF, and the friction torque ($\tau_F$) predicted by the PINNs. The desired motor currents $i_d$ are derived from the motor and gearbox dynamics (Eq.~\ref{eq:motor_gearbox_dynamics}), assuming $J_m \ddot{\theta} \approx 0$.}
    \vspace{-14pt}
    \label{fig:controlarchitecture}
\end{figure*}

\subsection{UKF for Joint Torque Estimation}
\label{subsec:ukf}
The Unscented Kalman Filter (UKF) plays a crucial role in our framework by providing the estimation of joint torques, essential for sensorless torque control of humanoid robots. The UKF is particularly well-suited for this task due to its ability to handle nonlinear system dynamics and its robustness to noise and uncertainties. Our UKF formulation incorporates several key innovations and enhancements that distinguish it from the previous approach~\cite{sorrentino_ukf-based_2024} designed for fixed-base systems. A significant innovation is the integration of PINN-based friction estimates ($\tau_F$) as a direct measurement input to the UKF. Unlike the previous approach that treats friction as an unmodeled disturbance or attempts to estimate it as part of the system state, we leverage the PINN's ability to provide explicit estimates of friction torque. The estimates are incorporated into the UKF's measurement update step, allowing the filter to account for and compensate for friction effects accurately. The UKF's process model has been refined to reflect the system dynamics and the influence of friction accurately. Instead, we move beyond simplified friction models and leverage the PINN's predictive capabilities to inform the process model. In addition, we have adapted the UKF formulation to be suitable for floating-base systems. This involved a non-trivial modification of the state vector and the system model to incorporate accelerometer and gyroscope measurements, which are essential for setting the robot base state.
% \textcolor{red}{This work extends the formulation to floating-base dynamics, requiring modifications in the definition of process and measurement dynamic models. Specifically, we integrate base acceleration and angular velocity directly from the robot's accelerometer and gyroscope into the state vector, enabling real-time adaptation to dynamic conditions. Another key innovation of our UKF formulation is incorporating PINN predictions as direct measurements rather than treating friction torques solely as estimated disturbances~\cite{sorrentino_ukf-based_2024}. Friction is now explicitly represented in the process and measurement dynamic equations as an observable state. These improvements significantly enhance the accuracy and robustness of joint torque estimation.}
The state and measurement vectors include additional states and measurements compared to previous UKF implementations:
\begin{equation}
    \begin{aligned}
    &x_k = \begin{bmatrix}\hat{\dot s}^\top & \hat{\tau}_{m}^\top & \hat{\tau}_{F}^\top & \hat{f}_{FT}^\top  & \hat{f}_{ext}^\top & \hat{\alpha}_{acc}^\top & \hat{\omega}_{gyro}^\top \end{bmatrix} ^\top_k \; , \\
    &y_k = \begin{bmatrix}\dot{s}^\top & I_{m}^\top & \tau_F^\top & f_{FT}^\top & \alpha_{acc}^\top & \omega_{gyro}^\top \end{bmatrix}^\top_k \; ,
    \end{aligned}
    \vspace{-2pt}
\end{equation}
being $\hat{\dot s}_k$ the joint velocities, $\hat{\tau}_{m,k}$ the motor torques, $\hat{\tau}_{F,k}$ the friction torques, $\hat{f}_{FT,k}$ the vector of the external wrenches applied to the FT sensors, $\hat{f}_{ext,k}$ the estimated unknown wrench, $\hat{\alpha}_{acc,k}$ the vector of estimated base accelerations, and $\hat{\omega}_{gyro,k}$ the vector of estimated base angular velocities. The measurement vector $y_k$ includes joint velocities $\dot{s}_k$, motor currents $i_{m,k}$, friction torques $\tau_{F,k}$, FT sensor readings $f_{FT,k}$, accelerometer linear accelerations $\alpha_{acc,k}$, gyroscope angular velocities $\omega_{gyro,k}$. The input vector $u_k$ includes the joint position vector $s_k$. The $k$ subscript denotes the time step.

The process model predicts the evolution of the state vector based on the dynamic equations detailed below.
\begin{itemize}
    \item[-] \textit{Joint velocity dynamics}: derived from Eq.~\ref{eq:dynamics_sensor_frame}:
    \begin{equation}
    \small
        \begin{aligned}
        &\ddot{s}{=} M_s(q)^{-1} [ \hat{\tau}_m {-} \hat{\tau}_F {-} C(q,{}^B \bar{\nu}_{A,B}) {-} M_{sb}(q) {}^B \alpha_{A,B}^g \\ &{+} \sum_{k} {J_{FT}^{k^\top}(q) \hat{f}_{FT}^k} {+} J_{ext}^\top(q) \hat{f}_{ext} ] \; .
        \end{aligned}
        \vspace{-2pt}
    \end{equation}
    The base acceleration and angular velocity measurements are provided by the state variables $\hat{\alpha}_{acc,k}$ and $\hat{\omega}_{gyro,k}$, respectively. The linear acceleration ${}^S \dot{v}^g_{A,S}$ and the angular velocity ${}^S \omega^g_{A,S}$ of the base in the sensor frame $S$, and are converted in the base frame $B$ as:
    \begin{equation}
        \begin{aligned}
        &{}^B \omega^g_{A,B} {=} {}^B R_S {}^S \omega^g_{A,S} \\
        &{}^B \alpha^g_{A,B} {=} {}^B R_S {}^S \dot{v}^g_{A,S} {-} {}^B \omega^g_{A,B} {\times} ({}^B \omega^g_{A,B} {\times} {}^B o_{B,S}) ,
        \end{aligned}
    \end{equation}
    where ${}^B R_S$ is the rotation matrix from the sensor frame $S$ to the base frame $B$, and ${}^B o_{B,S}$ is the position of the sensor frame $S$ in the base frame $B$. The angular base acceleration is approximated to zero.
    \item[-] \textit{Motor torques, Friction torques, FT sensors, contact wrenches, IMU dynamics}: assumed constant over short time intervals:
    \begin{equation}
    \dot{\hat{\tau}}_{m} = \dot{\hat{\tau}}_{F} = \dot{\hat{f}}_{FT} = \dot{\hat{f}}_{ext} = \dot{\hat{\alpha}}_{acc} = \dot{\hat{\omega}}_{gyro} = 0 \; .
    \vspace{-2pt}
    \end{equation}
\end{itemize}

The measurement model is extended to include direct observations of friction torque from the PINN predictions:
\begin{itemize}
\item[-] \textit{Joint velocity}: $\dot{s} = \hat{\dot s}$.
\item[-] \textit{Motor current}: $I_{m} = \frac{\hat{\tau}_{m}}{R k_t}$.
\item[-] \textit{Friction torque}: $\tau_{F} = \hat{\tau}_{F,PINN}$.
\item[-] \textit{Force/Torque sensors}: $f_{FT} = \hat{f}_{FT}$.
\item[-] \textit{Base linear acceleration}: $\alpha_{acc} = \hat{\alpha}_{acc}$
\item[-] \textit{Base angular velocity}: $\omega_{gyro} = \hat{\omega}_{gyro}$.
\end{itemize}

\section{Experimental Validation on Balancing Task}
\label{sec:results}

% Aggiungi la parte del KF
This section presents the experimental validation of the proposed torque control framework on the ergoCub humanoid robot during a balancing task (Figure~\ref{fig:ergocub}). This task provides a scenario for evaluating torque control strategies requiring precise force regulation, disturbance rejection, and dynamic adaptation. We control $18$ joints and use measurements from feet, waist, arm IMUs, and feet FTs, besides all the motor current sensors and joint and motor encoders. The control system, implemented in C++, runs on a 10th-gen Intel Core i7 laptop with Ubuntu Linux 22.04. KF gains are optimized using a GA with PyGAD~\cite{Gad2023PyGAD}, running for 50 generations with 60 parents selected for mating. The population size is 120, using $k$-tournament selection ($k=4$) and a two-point crossover strategy. A 20\% per gene mutation rate with random mutations is used, preserving the top 10\% of solutions in each generation. Data for the GA consist of a 6-minute trajectory sampled at 1 KHz. Optimizing the values takes around 30 minutes per joint on an AMD EPYC 7513 32-core Processor. The source code and supplementary materials, including videos, are available online~\url{https://github.com/ami-iit/paper_sorrentino_ral2024_balancing_torque}.\looseness=-1

\subsection{Controller architecture}
\label{subsec:controllerarchitecture}
\begin{figure}[t]
    \centering
    \includegraphics[width=\linewidth]{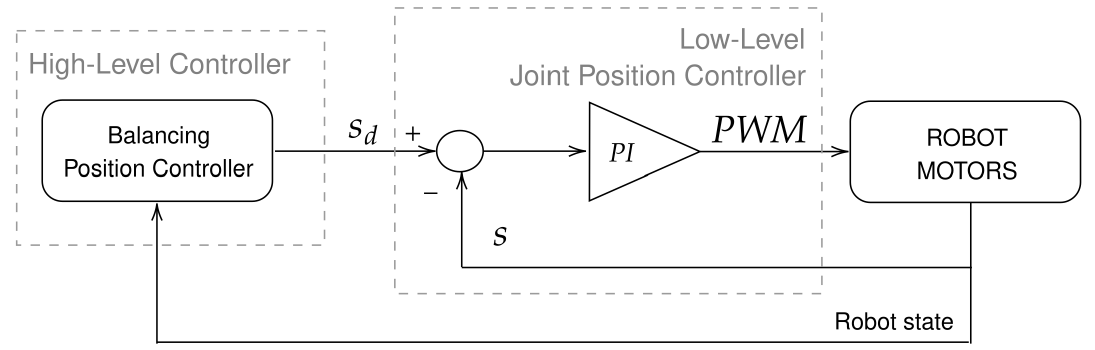}
    % \vspace{-12pt}
    \caption{Position control scheme. The balancing controller generates the desired joint positions, which are sent to the position controllers of each joint, which generate the PWM signals actuating the motors.}
    \label{fig:positioncontrolschema}
    \vspace{-18pt}
\end{figure}

The framework integrates a high-level momentum-based controller~\cite{nava_stability_nodate} with a low-level joint torque controller. The high-level controller computes desired joint torques by regulating the robot's centroidal momentum at 100 Hz. The low-level PI controller refines torque commands at 1 kHz using feedforward torque commands from the high-level controller, torque feedback from either RNEA or the proposed UKF-based estimation, and friction compensation through a PINN-based estimator. Finally, the joints are controlled through motor currents regulated at 20 KHz. Figure~\ref{fig:controlarchitecture} illustrates the overall control architecture, showcasing the interaction between high-level and low-level controllers and the integration of PINN and UKF components. Additionally, we compare balancing experiments performed through a position controller (Figure~\ref{fig:positioncontrolschema}), highlighting the advantages of achieving compliance through torque control.

\begin{figure*}[h]
    \centering
    \vspace{2pt}
    \begin{subfigure}[b]{0.48\textwidth}
        \centering
        \includegraphics[width=0.98\textwidth]{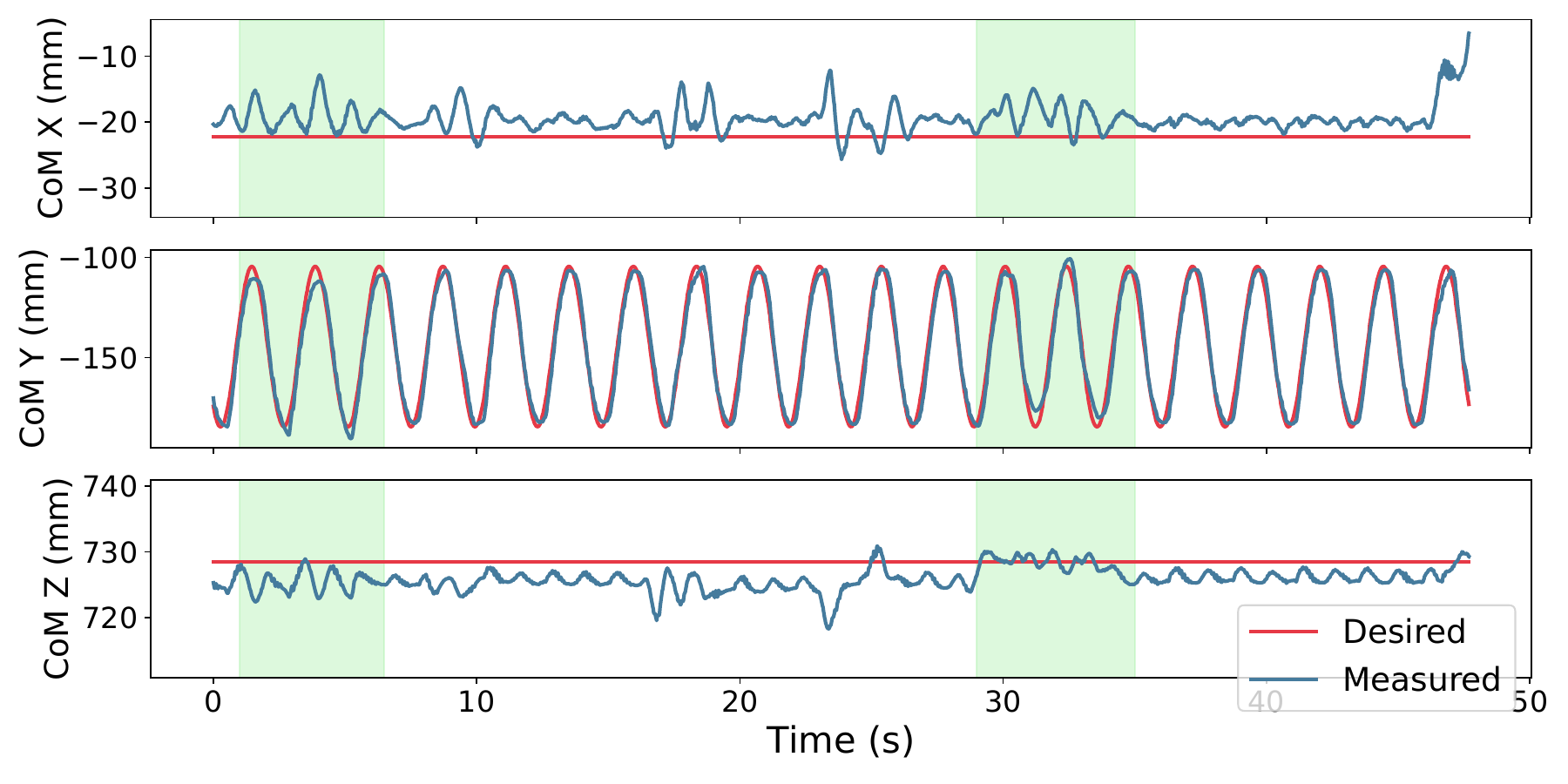}
        \caption{CoM tracking with RNEA-PINN.}
        \label{fig:rneapinn_com}
    \end{subfigure}
    \begin{subfigure}[b]{0.48\textwidth}
        \centering
        \includegraphics[width=0.98\textwidth]{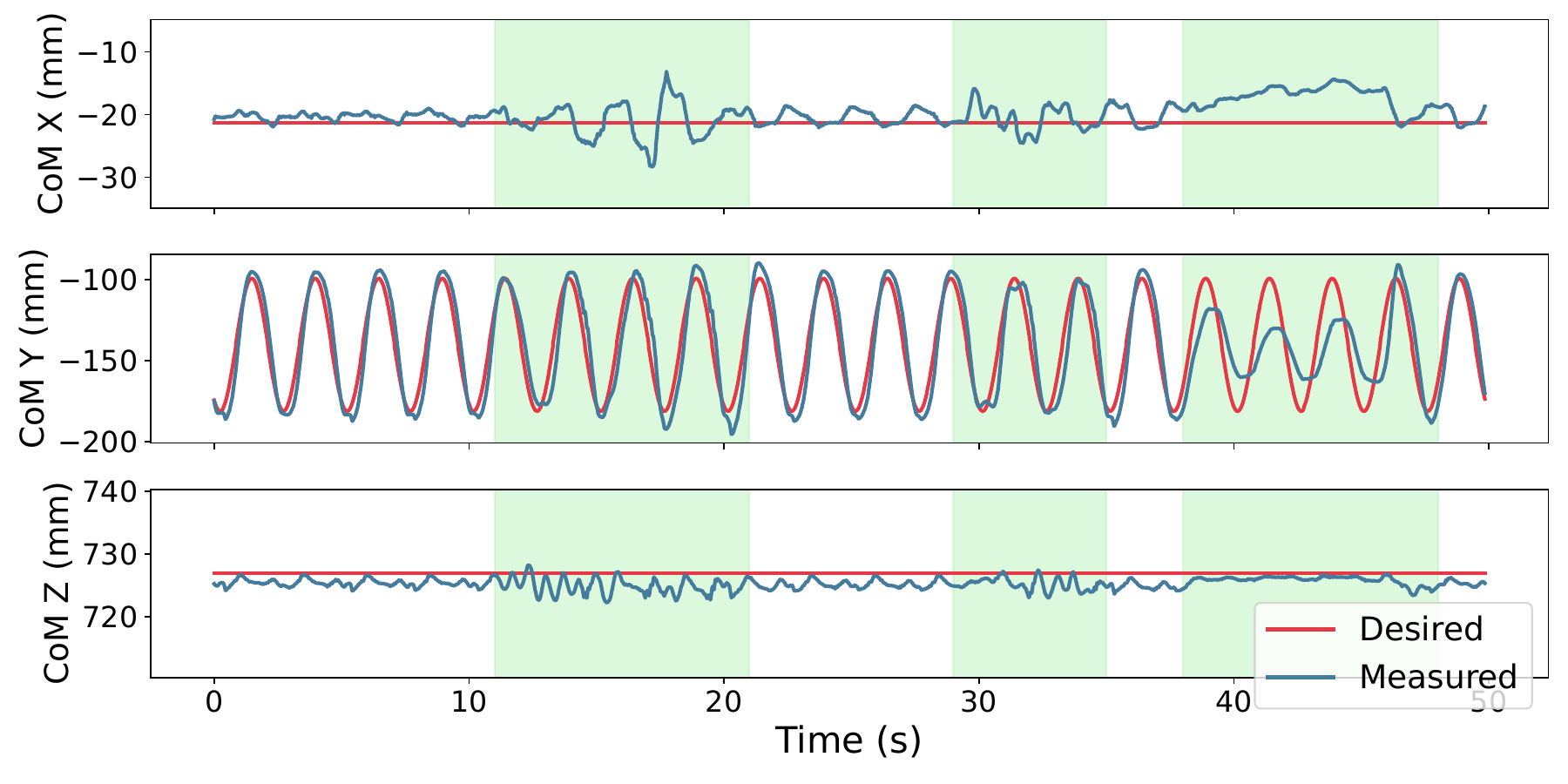}
        \caption{CoM tracking with UKF-PINN.}
        \label{fig:ukfpinn_com}
    \end{subfigure}
    \caption{CoM tracking comparison: \textit{RNEA-PINN} (left) vs. \textit{UKF-PINN} (right). Green rectangles indicate external contacts.}
    \label{fig:rneapinn_ukfpinn_com}
    \vspace{-15pt}
\end{figure*}

\subsection{Validation Methodology}
\label{subsec:validationmethodology}
We evaluate six control configurations, combining feedforward, torque feedback (RNEA or UKF), and PINN-based friction compensation:
\begin{enumerate}
\item \textit{Feedforward}: no torque feedback and friction compensation.
\item \textit{RNEA No Compensation}: torque feedback using the RNEA, no friction compensation.
\item \textit{UKF No Compensation}: torque feedback using the UKF, no friction compensation.
\item \textit{Feedforward PINN}: feedforward control with PINN-based friction compensation.
\item \textit{RNEA-PINN}: torque feedback via RNEA with friction compensation.
\item \textit{UKF-PINN}: fully integrated approach combining UKF with PINN-based friction compensation.
\end{enumerate}

Experiments include: (i) CoM tracking along a sinusoidal trajectory, (ii) external disturbances, and (iii) adaptation to environmental changes, introducing/removing an object under one foot. Controller gains are held constant across all configurations to ensure fair comparisons and isolate torque estimation and friction compensation contributions.

In addition, We conduct experiments to compare balancing in torque mode versus position mode under dynamic environmental changes. Specifically, we introduce an object under one foot and remove it while evaluating the robot's ability to maintain balance. The robot is expected to fail in position control due to its rigidity, whereas it should adapt to environmental changes in torque control.
Performance metrics included:
\begin{itemize}
\item[-] Center-of-Mass (CoM) and torque tracking Accuracy.
\item[-] Stability and compliance, involving response to external disturbances and non-modelled environmental changes.
\item[-] Energy efficiency evaluated through joint torque magnitudes. 
\item[-] Scalability and reproducibility.
\end{itemize}

\subsection{Results}\label{subsec:results}
The results of the experimental evaluation of the proposed framework on the ergoCub humanoid robot are presented in this section. These results include a comparative analysis of the six control configurations described in \ref{subsec:validationmethodology} and a comparison with position control, with key findings supported by accompanying figures and a video submission.

\subsubsection{Center-of-Mass and Torque Tracking}

\begin{figure*}[h]
    \centering
    \vspace{2pt}
    \begin{subfigure}[b]{\textwidth}
        \centering
        \includegraphics[width=\textwidth]{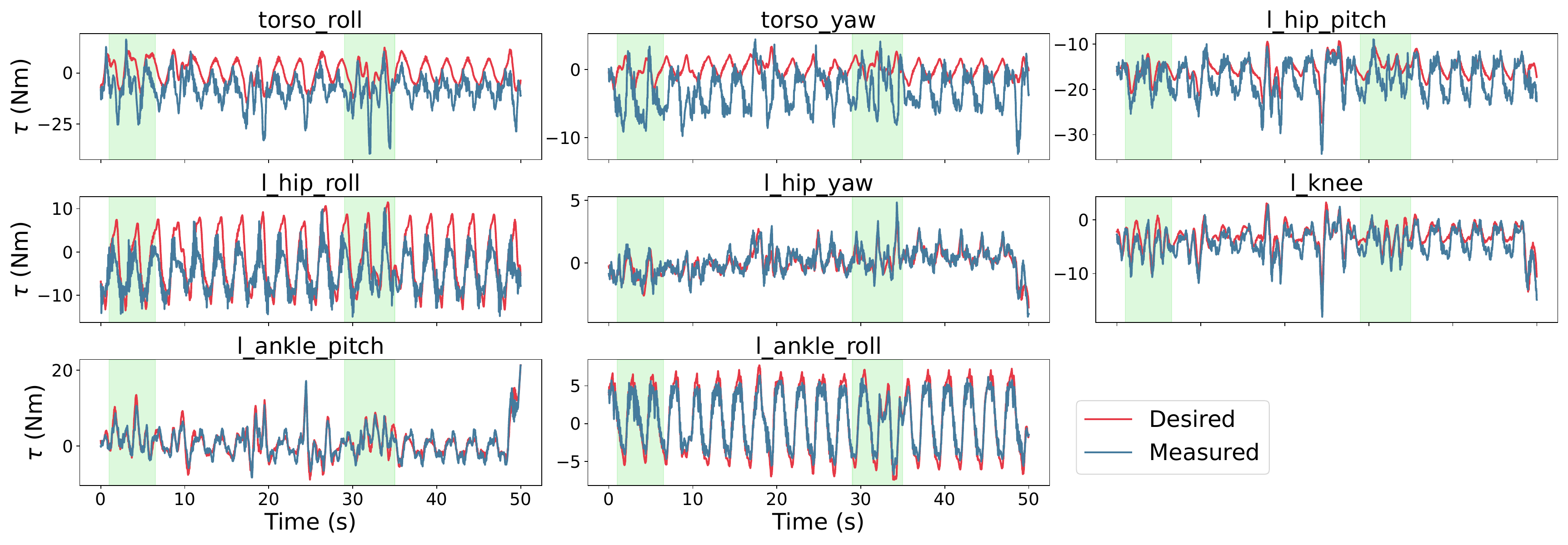}
        \caption{Torque tracking with RNEA-PINN.}
        \label{fig:rneapinn_trq}
    \end{subfigure}
    \begin{subfigure}[b]{\textwidth}
        \centering
        \includegraphics[width=\textwidth]{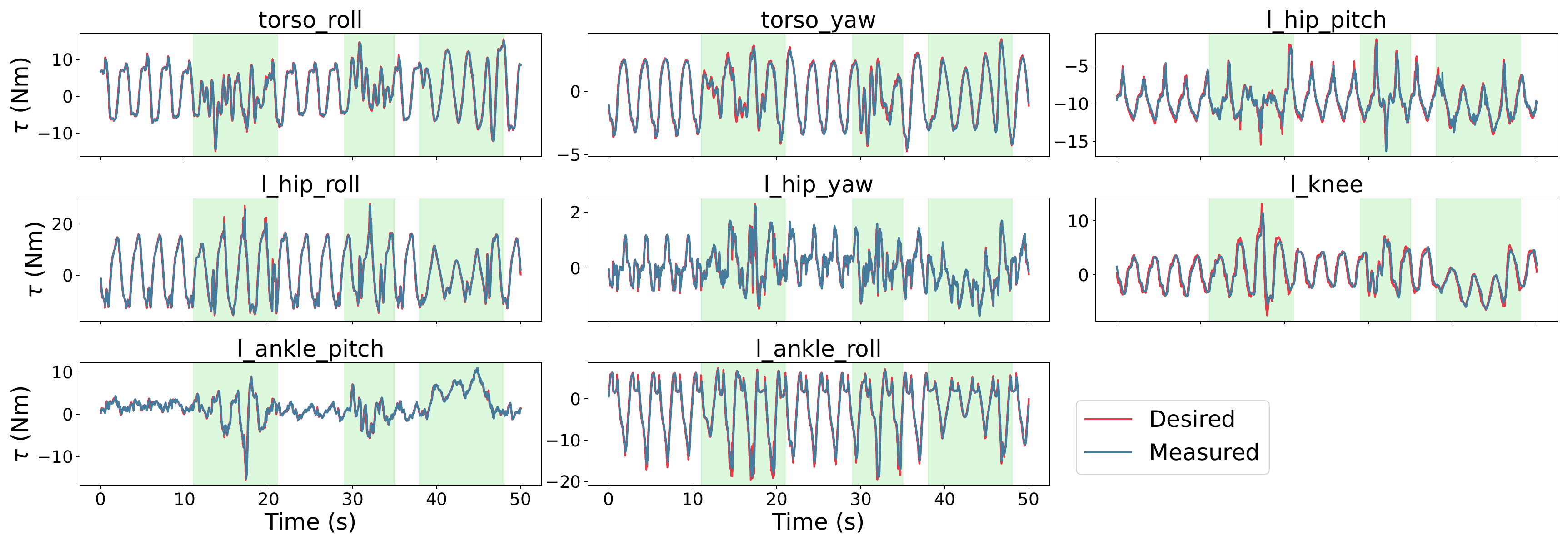}
        \caption{Torque tracking with UKF-PINN.}
        \label{fig:ukfpinn_trq}
    \end{subfigure}
    \caption{Torque tracking comparison. Green rectangles highlight time intervals where external contacts are applied.}
    \label{fig:rneapinn_ukfpinn_trq}
    \vspace{-14pt}
\end{figure*}

We begin by analyzing the CoM and torque tracking performance across the six control configurations. The CoM tracking performance reveals a balance between tracking accuracy and system stability across different conditions. Table~\ref{tab:com_errors_before} presents the mean and maximum CoM errors before external disturbances, while Table~\ref{tab:torque_tracking} quantifies the torque tracking performance.
As expected, feedforward configurations exhibit the highest torque errors due to the absence of feedback. Focusing on the closed-loop configurations without friction compensation, \textit{UKF No Compensation} outperforms \textit{RNEA No Compensation} (RMSE: 2.03 Nm vs. 4.37 Nm), demonstrating the benefits of the UKF's sensor fusion.
\textit{RNEA-PINN} achieves the lowest CoM errors (0.0028 m in z, 0.0038 m in y) under nominal conditions. However, unmodeled disturbances cause significant torque tracking errors (Figure~\ref{fig:rneapinn_trq}), leading to CoM divergence after repeated contacts, as illustrated in Figure~\ref{fig:rneapinn_com}. This divergence is attributed to the RNEA estimation's inability to detect unmeasured interactions during external contacts. The green rectangles in Fig. \ref{fig:rneapinn_ukfpinn_com} and \ref{fig:rneapinn_ukfpinn_trq} highlight the periods of external contact.

On the other hand, \textit{UKF-PINN} exhibits slightly higher CoM errors in nominal conditions but demonstrates superior torque tracking (RMSE: 1.08 Nm, MAE: 0.70 Nm) and effectively adapts to external forces, maintaining stability. As depicted in Figures~\ref{fig:ukfpinn_com} and~\ref{fig:ukfpinn_trq}, \textit{UKF-PINN} maintains good CoM and torque tracking even under multiple random disturbances, which introduce significant variations in the robot's state, thus constituting a dynamic environment. The UKF effectively detects changes in joint torques when external contacts occur, enhancing the system's responsiveness.

\begin{figure}[t]
    \centering
    % \vspace{4pt}
    \includegraphics[width=\linewidth]{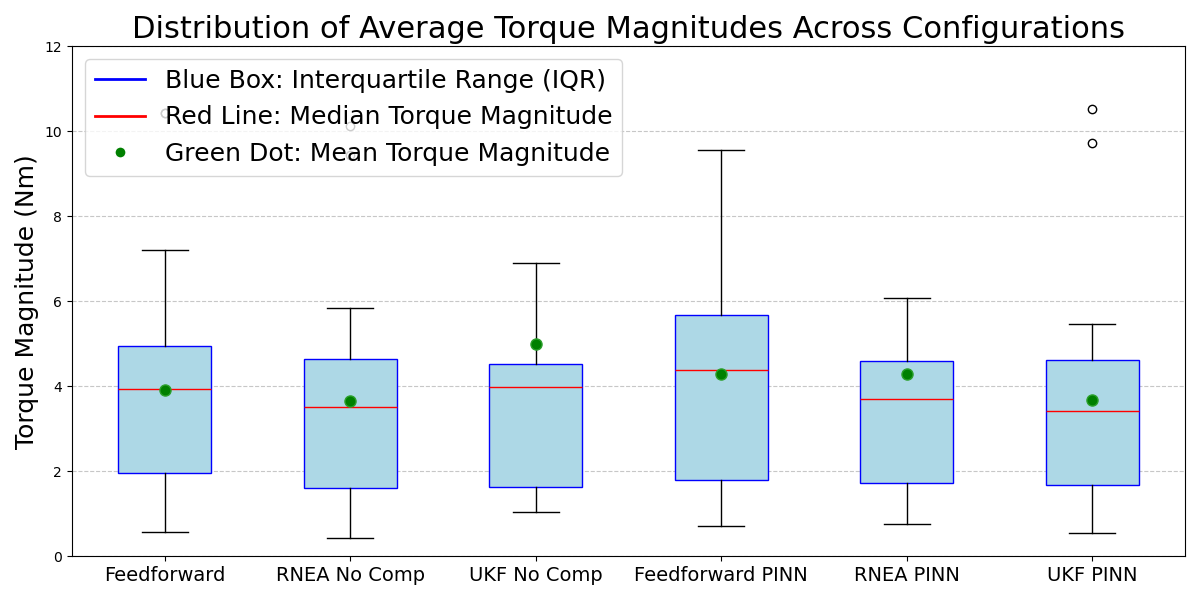}
    % \vspace{-12pt}
    \caption{Distribution of average torque magnitudes across configurations. The \textit{UKF-PINN} configuration achieves the lowest torque magnitudes.}
    \label{fig:average_torque}
    \vspace{-20pt}
\end{figure}

The advantages of \textit{UKF-PINN} extend to energy efficiency. As shown in Figure~\ref{fig:average_torque}, this configuration achieves the lowest average torque values, indicating smoother and more controlled torque profiles than other configurations. Other configurations, particularly feedforward approaches, exhibit higher torque efforts due to their inability to adapt dynamically to changes in system dynamics or external forces.

\begin{table}[t]
    \centering
    \caption{Mean and Maximum CoM Errors Before Disturbances (m)}
    \label{tab:com_errors_before}
    \resizebox{\columnwidth}{!}
    {
        \begin{tabular}{@{}lcc@{}}
            \toprule
            \textbf{Control Architecture} & \textbf{Mean Error (mm)} & \textbf{Max Error (mm)} \\ \midrule
            \rowcolor{red!25} % Highlight row with highest error
            Feedforward                 & [1.2, 13.1, 1.9] & [3.1, 24.2, 4.6] \\ \hline
            RNEA No Compensation         & [0.8, 6.5, 0.7] & [2.3, 15.2, 2] \\ \hline
            UKF No Compensation         & [3.2, 11.5, 6.4] & [7.2, 20.6, 7.5] \\ \hline
            Feedforward PINN            & [2.8, 9.4, 1.1] & [4.4, 25.4, 2.5] \\ \hline
            \rowcolor{green!25} % Highlight row with lowest error
            RNEA-PINN                    & [2.8, 3.8, 3.8] & [8.2, 18.1, 6.7] \\ \hline
            \rowcolor{green!25} % Highlight row with lowest error
            UKF-PINN                    & [2.9, 8.1, 0.6] & [4.5, 17.5, 1.3] \\ \bottomrule
        \end{tabular}
    }
    \vspace{-20pt}
\end{table}

\begin{table}[ht]
    \centering
    \caption{Torque Tracking Performance Metrics (Nm)}
    \label{tab:torque_tracking}
    \adjustbox{max width=\textwidth}{
    \begin{tabular}{@{}lccc@{}}
        \toprule
        \textbf{Control Architecture} & \textbf{MSE} & \textbf{RMSE} & \textbf{MAE} \\ \midrule
        RNEA No Compensation & 19.2 & 4.37 & 2.82 \\ \hline
        UKF No Compensation & 4.13 & 2.03 & 1.45 \\ \hline
        RNEA-PINN & 49.43 & 7.03 & 4.12 \\ \hline
        \rowcolor{green!25} % Highlight best result
        UKF-PINN & 1.18 & 1.08 & 0.7 \\
        \bottomrule
    \end{tabular}}
    \vspace{-18pt}
\end{table}

\begin{figure}[t]
    \centering
    \includegraphics[width=1\linewidth]{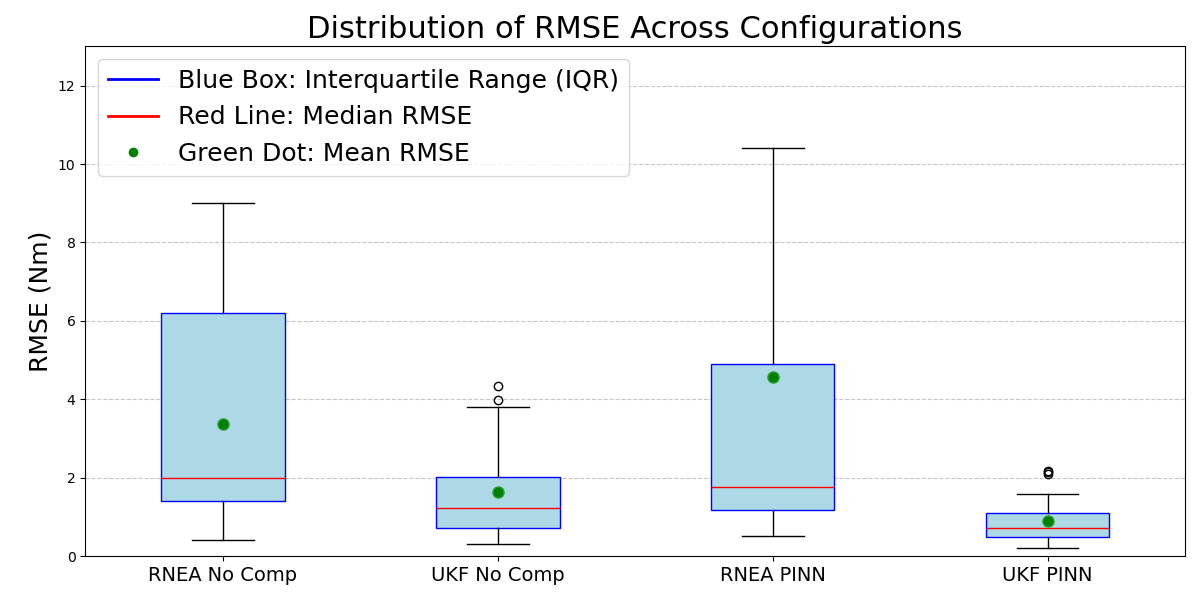}
    % \vspace{-12pt}
    \caption{Distribution of RMSE values for torque tracking across configurations. The \textit{UKF-PINN} configuration achieves the lowest RMSE and variance.}
    \label{fig:rmse_torque}
    \vspace{-20pt}
\end{figure}

\subsubsection{Stability and Compliance}
\label{subsubsec:stability}
Stability tests highlight the contrasting behaviors of \textit{RNEA-PINN} and \textit{UKF-PINN}. \textit{RNEA-PINN} demonstrates instability during prolonged experiments. This instability arises from its deterministic nature and reliance on model-based estimation, which struggles to adapt to unmodeled dynamics or unmeasured external disturbances. The result is often uncontrolled leg rotations, necessitating early termination of experiments. In contrast, \textit{UKF-PINN} maintains stability, even under stronger and more varied disturbances. This stability is achieved by leveraging sensor fusion and probabilistic estimation. The UKF's ability to combine information from multiple sources, including motor currents, IMUs, and PINN-based friction estimates, enables dynamic adaptation to environmental changes. This adaptability facilitates accurate torque tracking and efficient disturbance rejection, ensuring system recovery without failure and promoting compliant behavior.

\begin{figure*}[h]
    \centering
    \vspace{2pt}
    \begin{subfigure}[b]{\textwidth}
        \centering
        \includegraphics[width=\textwidth]{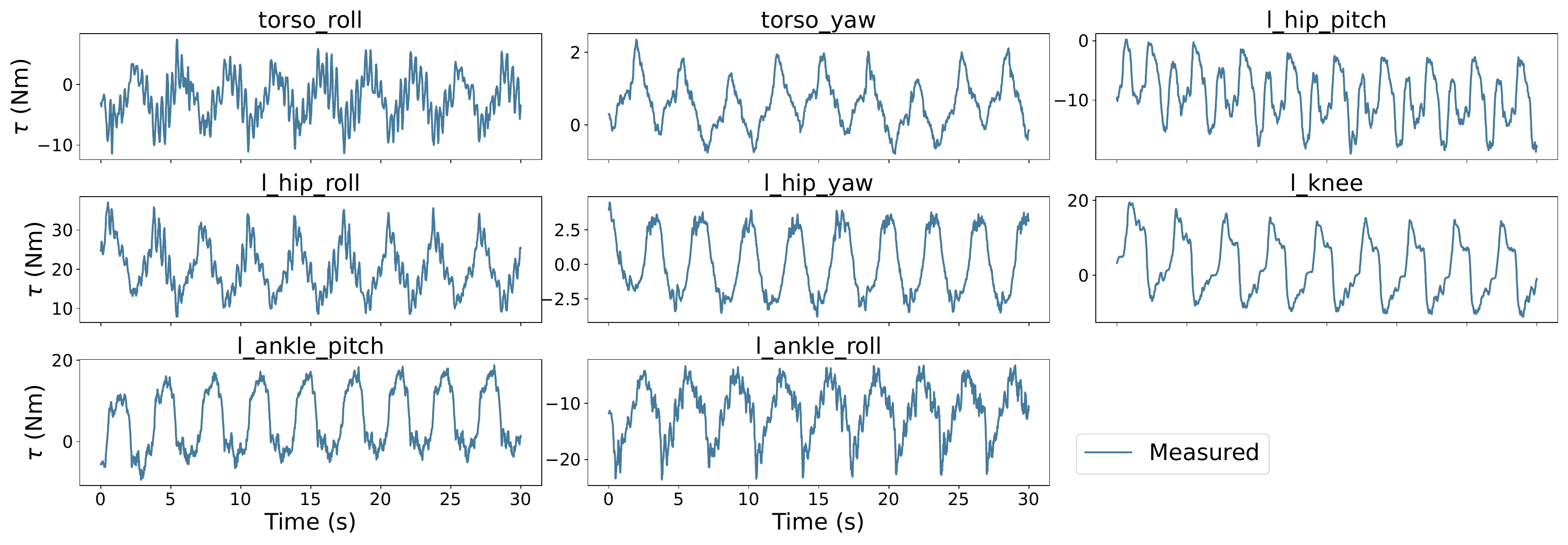}
        \caption{Joint torques required to balance on an object in position control mode.}
        \label{fig:object_position_torque}
    \end{subfigure}
    \begin{subfigure}[b]{\textwidth}
        \centering
        \includegraphics[width=\textwidth]{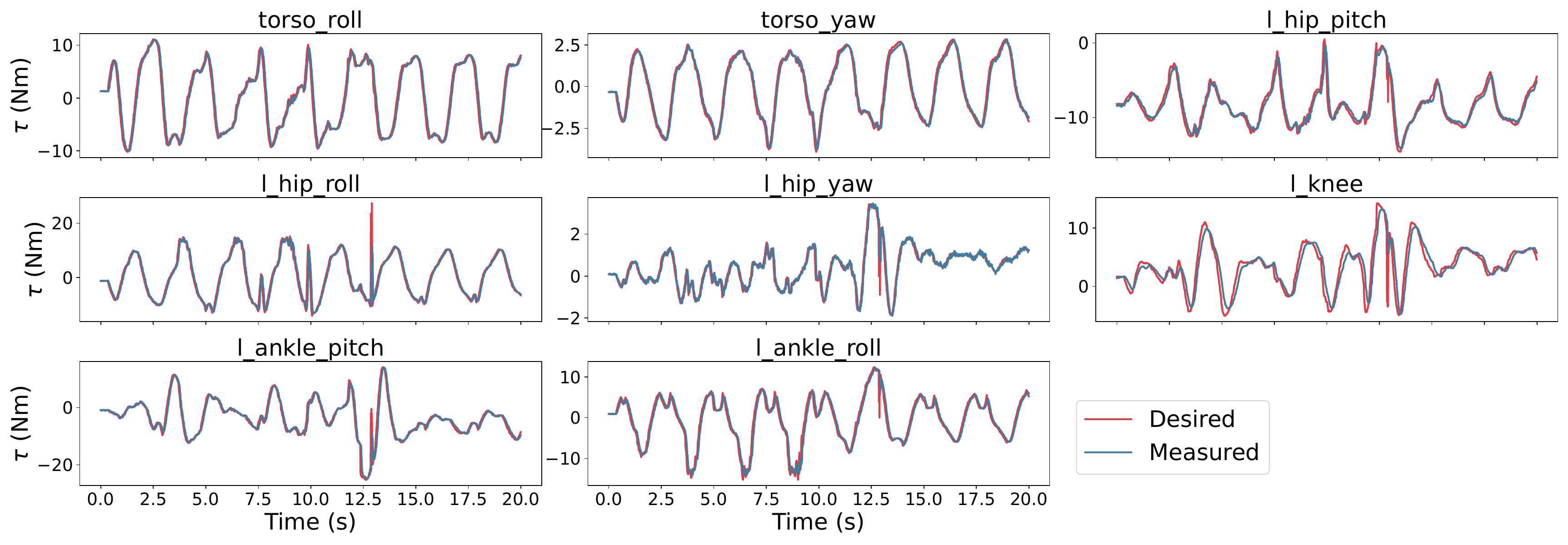}
        \caption{Tracking of joint torques while balancing on an object dynamically removed under the robot foot with the joints torque controlled.}
        \label{fig:object_torque_tracking}
    \end{subfigure}
    \caption{Comparison of joint torque magnitudes between position control and torque control during the object removal experiment. Torque control demonstrates significantly lower torque requirements, indicating improved energy efficiency and reduced stress on the robot's motors compared to position control.}
    \label{fig:object_trq}
    \vspace{-15pt}
\end{figure*}

To further illustrate the role of compliance, the robot's balancing behavior is evaluated under both position control and torque control strategies. This experiment challenged the robot's dynamic balance by testing its response to an unexpected loss of support. Initially, the robot balances with an object under its right foot, creating an unmodeled support condition. The object is then rapidly removed, forcing the robot to adapt to the abrupt change and maintain its balance. The experiments are detailed in the accompanying video. In particular, the experiment is repeated four times with different objects and placements to show the approach's robustness. The robot exhibits a rigid response in position control mode and fails to adapt to environmental change. This lack of compliance leads to an immediate loss of balance. In addition, as shown in Figure~\ref{fig:object_trq}, the required torques under position control are significantly higher than those under torque control. Higher torques demand more electrical current, impacting energy consumption and imposing greater mechanical stress on the motors and their components. In contrast, the lower torque magnitudes achieved by torque control enhance energy efficiency and minimize mechanical stress on the motors. The torque controller effectively modulates the output torques to adapt to the changing support conditions, preventing sudden loss of balance. This improved stability, enabled by torque control on a sensorless humanoid robot thanks to the estimation techniques presented here, allows for superior disturbance rejection and adaptability compared to position control.

\subsubsection{Scalability}
\label{subsubsec:scalability}
Scalability tests confirm the robustness and portability of the \textit{UKF-PINN} framework. The framework's generalizability was evaluated on two ergoCub robots. Despite sharing the same design, these robots exhibit differences due to aging and usage. One of the two robots has increased mechanical wear compared to the newer unit, and frictional properties vary between joints due to manufacturing variations and factors like grease depletion and environmental conditions. These discrepancies pose challenges for model-based controllers, typically requiring re-identifying system parameters for each hardware instance. However, the \textit{UKF-PINN} framework successfully replicates the balancing task on both robots without re-tuning, demonstrating its ability to handle friction and mechanical properties variations. This result underscores the robustness and generalizability of \textit{UKF-PINN}, highlighting its potential for deployment across multiple similar robots without extensive re-identification.

\section{Conclusions}
\label{sec:conclusions}

This paper proposes a novel framework for whole-body torque control of humanoid robots, integrating UKF-based joint torque estimation with PINN-based friction compensation.
The primary contributions include PINN-based friction modeling, which effectively learns nonlinear static and dynamic friction by leveraging joint and motor velocity readings, and UKF-based torque estimation, where the UKF is formulated for floating-base humanoid robots, utilizing PINN-based friction estimates as direct measurement, significantly improving torque estimation robustness compared to traditional methods like RNEA. Experimental validation on the ergoCub humanoid robot demonstrates improved torque tracking accuracy, enhanced energy efficiency, superior disturbance rejection, and scalability.

Potential limitations of the framework include the computational cost associated with the UKF. Future work will investigate methods to optimize the computational efficiency of the framework. Future developments will focus on further validation in more complex scenarios, the active development of a walking pipeline to test the proposed torque control framework in a full locomotion task, and the application of this framework to other humanoid robots.

%%%%%%%%%%%%%%%%%%%%%%%%%%%%%%%%%%%%%%%%%%%%%%%%%%%%%%%%%%%%%%%%%%%%%%%%%%%%%%%%

\bibliography{references}
\bibliographystyle{IEEEtran}

\end{document}